# Neural Machine Translation between Herbal Prescriptions and Diseases

## Applications of convolutional and recurrent neural networks to herbal prescription big data

Sun-Chong Wang
Institute of Systems Biology and Bioinformatics
Department of Biomedical Sciences and Engineering
National Central University
Chungli, Taoyuan 32001, Taiwan
scwang@ncu.edu.tw

***Abstract*—Animals eat bitter tasting leaves when they get infested, just as humans sought medicinal plants when they felt uncomfortable. As people gained more experiences, formulas were made from brewing a concoction of multiple herbs at predetermined proportions, targeting disorders. Herbs and formulas are now available in the market in the form of concentrated extract powders, and an herbal prescription is now prescribed as a combination of many granulated formulas and herbs at a defined proportion, addressing complex diseases such as cancer. Modern herbalism is in a sense more of an art than a science, and mastering the art of herbalism presents a challenging task. Recent progress in artificial neural networks has reached a range of milestones from pattern recognition, language translation to strategic game championship. The current study applies deep learning to herbalism. Toward the goal, we acquired the de-identified health insurance reimbursements that were claimed in a 10-year period from 2004 to 2013 in the National Health Insurance Database of Taiwan, the total number of reimbursement records equaling 340 millions. Information in each record includes patient's sex, year of birth, primary disease, secondary disease, tertiary disease, and herbal prescription including dose, daily frequency, and number of days, and date of prescription. Two artificial intelligence techniques were applied to the dataset: 1) residual convolutional neural network multitask classifier (rCNN) and 2) attention-based recurrent neural network (aRNN). The former works to translate from herbal prescriptions to diseases; and the latter from diseases to herbal prescriptions. Analysis of the classification results of rCNN indicates that herbal prescriptions are specific to: anatomy, pathophysiology, sex and age of the patient, and season and year of the prescription. Further analysis identifies temperature and gross domestic product as the meteorological and socioeconomic factors that are associated with herbal prescriptions. Analysis of the neural machine transitional result indicates that the aRNN learnt not only syntax but also semantics of diseases and herbal prescriptions.**

### I. Introduction

The last decade has witnessed great advances in artificial neural network technologies [1–3]. This wave of artificial intelligence (AI) renaissance can be attributed to three converging developments: 1) algorithmic breakthroughs; 2) powerful computations; and 3) big data. Applications of AI have since spread from image classification, voice recognition and language translation to self-driving vehicles, personal virtual assistants and investment portfolio management. This study of AI on herbalism represents an attempt to expand the list of AI applications.

The numbers of tunable parameters in artificial neural networks are typically much greater than those of the data points. The outnumbering is meant for AI to make generalizations. Good predictions however rely on training data of high quality. To this end, we obtained the herbal prescriptions and the diseases the prescriptions were used to treat from the insurance reimbursement records in the National Health Insurance Database, Taiwan [4]. The dates of the records span a decade from 2004 to 2013 and the number of paired herbal prescriptions and disease diagnoses totals 340 millions.

The prescribed herbal prescriptions in the reimbursement dataset are in the form of concentrated herbal extract granules. This form of herbal prescriptions, manufactured by GMP-certified pharmaceutical companies, is believed to be better standardized. It is also convenient for consumers in the mobile era, compared to the liquid form of traditional Chinese medicine (TCM) decoctions. The second feature of the data is that diseases were diagnosed by western medicine and coded in the International Statistical Classification of Diseases version 9 (ICD-9 codes) [5]. This coding system, replacing TCM diagnostics, greatly helps those users who are not familiar with TCM terminology. Another feature is that not only prescription and diagnosis but also patient's sex, age, and date and dose of prescription are available in the dataset. Deep learning of state of the art AI methods on this unique dataset has the potential to achieve personalized herbal medicine that is realistic.

Previous studies characterizing usage patterns of granulated herbal prescriptions found that a typical herbal prescription consists of six components: two formulas and four herbs and that the proportion of the components by weight can be modeled by Zipf's law [6,7]. The current study confirms these. An herbal prescription is therefore conveniently expressed as a collection of 'weighted' tokens, where a token can be a formula or herb. Given this, two AI technologies have been developed for the analysis of the herbal big data: 1) residual convolutional neural network multitask classifier which classifies an herbal prescription into disease, sex, age of patient and month and year of prescription; and 2) an attention-based recurrent neural network which translates disease, sex, age of patient and season into herbal prescription. The key to the first approach lies in converting herbal prescriptions into image data while the second in augmenting each output prescription by an extra token, i.e. Zipf's exponent, describing the weight distribution of the components in the prescription. Joint application of the two technologies achieves real-time two-way translation between herbal prescriptions and patient's phenotypes.

## II. Material and Methods

### A. National Health Insurance Database, Taiwan

The healthcare system of Taiwan, a single-payer compulsory social insurance program run by the government since 1995, covers both western medicine and TCM. Although most of the residents of Taiwan, a population of 23 millions, firstly seek help from western medicine, about 30% of them routinely patronize TCM. Training of TCM doctors in Taiwan includes both modern western medicine and TCM courses. So, Taiwanese TCM professionals are proficient in western medicine. The reimbursement records recorded disease diagnoses in ICD-9 codes. In a record, there can be up to three ICD-9 codes for the primary, secondary and tertiary health problems of the patient. Information about the herbal prescriptions in the records includes weights in grams of the component formulas/herbs per serving, frequency of serving per day and length of serving in days. The number of unique formulas prescribed in the period between 2004 and 2013 is 303 and that of herbs is 415. Other information includes patient's sex, birth year and prescription date. Application for and access to the data was reviewed and approved by the Institutional Review Board on 22 July, 2015 (LHIRB IRB #: 15-014-C0, Landseed Hospital, Taiwan).

### B. Preprocessing of the reimbursement data

Image classification is one of the areas where AI demonstrates its prowess. To take advantage of this, we converted an herbal prescription into a one-dimensional (1-d) array of size 840 in the following way. The array elements were initialized to zero. The doses (weight in grams per serving) of the component formulas and herbs were stored in the first 718 (= 303 + 415) elements of the 1-d array. Formula 1 mapped to element 1, formula 2 to element 2, …, herb 1 to element 304, …, and so on. Each of the doses was normalized by dividing the weight by the maximum weight, which is usually 5 grams, when the formula or herb is served alone. Next, one of the array elements 719 to 723 was set to 1 if the TCM treatment included acupuncture. The five elements correspond to the five acupuncture treatment modalities. A one in elements 724 to 750 was used to code one of the 27 daily serving schedules (for example, three serves after meals, two serves after and before bedtime, …, etc). A one in elements 750 to 840 marked the duration of serving, from 1 day to 90 days. The values of the 1-d array elements were therefore bounded between 0 and 1, reminiscent to the pixel intensities of a one-channel image.

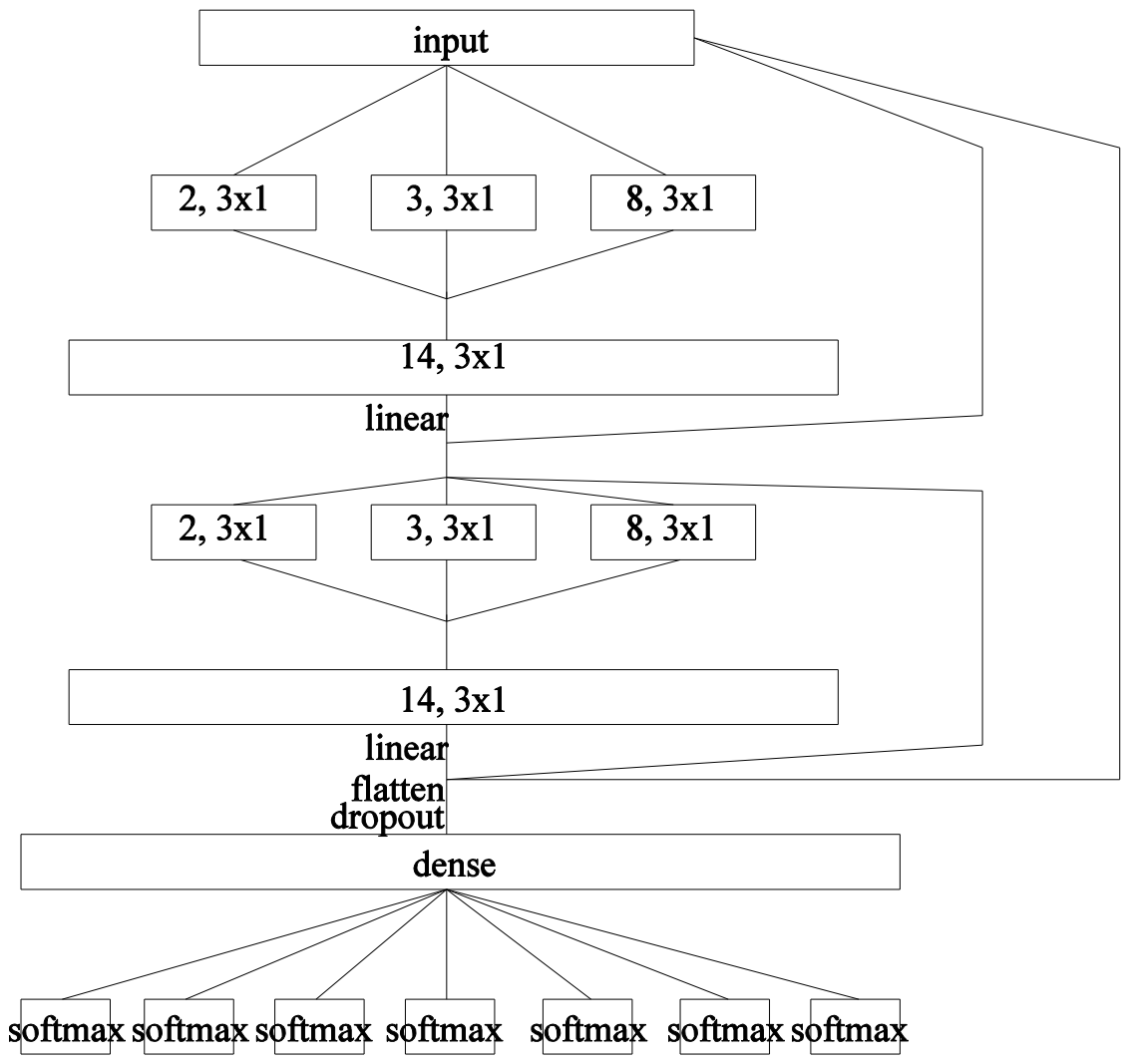

**Figure 1** A residual network multitask classifier consisting of cascading modules of convolutional neural networks. The activation is relu unless stated otherwise.

The frequency distribution of disease diagnoses in the dataset is not even. For example, most TCM patients suffered from acute nasopharyngitis (i.e. common cold, ICD-9 = 460). Furthermore, most TCM patients were middle-aged. There were more female TCM patients than male. To balance class sizes, we replaced the records with their medoids by the use of the *k*-medoid clustering algorithm [8]. This down sampling method, similar to the selection of senators to represent individual states, flattened the distribution of classes and reduced the number of records from 340 millions to 42,958,000. In subsequent learning, 42 millions of them were used for training and the rest for testing.

### C. Residual network classifier and spectral clustering

The state of the art convolutional neural networks are the so-called residual networks featuring short-cut connections between non-adjacent layers [9]. The residual network multi-task classifier (rCNN) developed in the current study is shown in Fig. 1. The design of two, three and eight filters, in parallel, in the first convolutional layer was inspired by the *qi-blood*, *three-burner* and *eight-guideline* diagnostic methods of TCM. Similarly, the choice of 14 filters in the second convolutional layer was inspired by the 12 + 2 meridians in TCM. The output layer has seven softmax functions for the seven classification tasks: primary disease, secondary disease, tertiary disease, sex, age, month and year. Their levels are, respectively, 909, 909, 909, 2, 105, 12 and 10. 909 arises from the number of unique 3-digit ICD-9 codes in the dataset, and 105 from the range of patient ages in the dataset.

Different herbal prescriptions can be prescribed to patients with the same ICD-9, just as images of different cats belong to the same cat label. Same herbal prescriptions can be prescribed to patients with different ICD-9s, just as cat images can be mistakenly classified into tiger class because cat and tiger resemble each other. To study the classification results, we applied the method of spectral clustering [10] on symmetrized confusion matrix obtained from the trained classifier on test data.

### D. Sequence to sequence learning and Zipf's exponent

Herbal prescription composing is oftentimes considered an art. With the vast amount of high quality herbal prescription data at hand, we are in a position to make AI master the art of herbalism. Toward the aim, we employed a 3-layer, 512-neuron per layer, recurrent neural network with attention mechanism (aRNN) [3], the other flagship of the AI fleet. If we liken patient's phenotypic information, i.e. the three ICD-9s, sex, age, plus the current month and year, to English, and herbal prescriptions to French, we are performing English to French translation. To apply neural machine translation, we ordered the component formulas/herbs in the herbal prescription according to their normalized weights. However, in addition to the order, two different prescriptions can have the same components but different proportions. For example, formula 1 + formula 2 in one prescription and formula 1 + 0.5 $\times$ formula 2 in the other prescription. In other words, our sentences have tones. To differentiate sentences with different tones, we modeled the weight distribution of each prescription by Zipf law to obtain a Zipf's exponent for each prescription. So, the first prescription above has a Zipf's exponent of 0 and the second 1. The distribution of the resulting Zipf's exponents was divided into 15 intervals, each of which was represented by a token. The number of our English tokens, i.e. vocabulary size, was 9,459, while that of French was 825. 9,459 comes mainly from the number of unique 5-digit ICD-9s in the dataset.

## III. RESULTS

There exist no previous results of herbal prescription classification prediction by algorithms. To evaluate the performance, we therefore compared the classification accuracy of the rCNN classifier with those of support vector machine (SVM) and *k*-nearest neighbors (*k*NN), which are known to be two of the best non-AI classifiers, on the same reimbursement dataset. To simplify matters, only herbal prescriptions for cancers (i.e. neoplasms) were tested, reducing the number of class labels to 94 cancers. The result in Table 1 shows that rCNN outperforms both SVM and *k*NN by a large margin.

TABLE I
PERFORMANCE AI AND NON-AI CLASSIFIERS

| Type | Classifier (optimal configuration) | Accuracy (%) |
|---|---|---|
| non-AI | SVM (kernel = radial, cost = 10, gamma = 0.5) | 32 |
| non-AI | *k*NN ($k$ = 1) | 29 |
| AI | rCNN (Figure 1) | 46 |

The rCNN clustering of cancers (the primary ICD-9) is shown in Fig. 2. Supplementary Fig. S1 shows the clustering result of all diseases. It is seen in Fig. 2 that cancers of, say, naso-, oro-, hypo-pharynx, pleura, gum, and other and unspecified parts of mouth congregate together, so do cancers of esophagus, stomach, colon, and rectum, indicating that diseases at nearby anatomic positions were treated by the same herbal prescriptions. Furthermore, benign neoplasms congregate. Similarly, the genitourinary disease cluster, musculoskeletal disease cluster, and injury and poisoning cluster in Supplementary Fig. S2-S4 indicates that disorders of the same physiological dyshomeostasis were treated the same way by the herbalists in Taiwan. As there are 18 categories of diseases (rf Fig. S1), in which cancers are one of them, according to ICD-9, if we calculate the distances between disease categories (from the selected eigenvectors of the unnormalized Laplacian of the symmetrized confusion matrix), the result in Fig. 3 shows that, in terms of the prescribed herbal prescriptions, cancers are closely related to digestive and immune disorders.

rCNN achieved an accuracy of 76% for patient sex classification from the herbal prescriptions. Furthermore, Fig. 4 and Supplementary Figs. S5 and S6 show how the predicted patient ages and prescription months and years by the rCNN congregate. The subclusters in the hierarchical clustering results indicate that patients' ages were measured in a unit of ~7 years and that herbal prescriptions, for the same disease, sex, age and year (month), could vary from season to season (from year to year).

The rCNN results above reveal that herbal prescriptions are specific to not only diseases, but also sex, age of the patient and the season and year the patient lives in. In the training of sequence to sequence translation by the aRNN, each of the source sequence therefore consisted of seven tokens: primary, secondary, tertiary ICD-9 codes, sex, age, season and year. The target sentence could have a variable number of tokens but the minimum is four: one for the Zipf's exponent, one for the formula (or herb), one for the frequency of serving per day, and one for the number of days. Once we have the Zipf's exponent, $z$, the weight distribution of the components is deciphered with the weight ratio going like: $1 : 1/2^z : 1/3^z : 1/4^z : \ldots$, and so forth, where 1, 2, 3, 4, …, are the order of the component in the output target sentence. Training was terminated after 2 epochs when the perplexity per word dropped to ~13. The performance of the aRNN on the test dataset achieved per word perplexities of 7, 12, 16, 21 for the target sentences of bucket sizes: 8, 10, 12, 18.

Vector representations of words are believed to be essential in the modeling of natural languages by artificial neural networks [11]. Figure 5 shows the words in the tSNE dimension reduction plot of the weights in the decoder embedding layer of the trained aRNN.

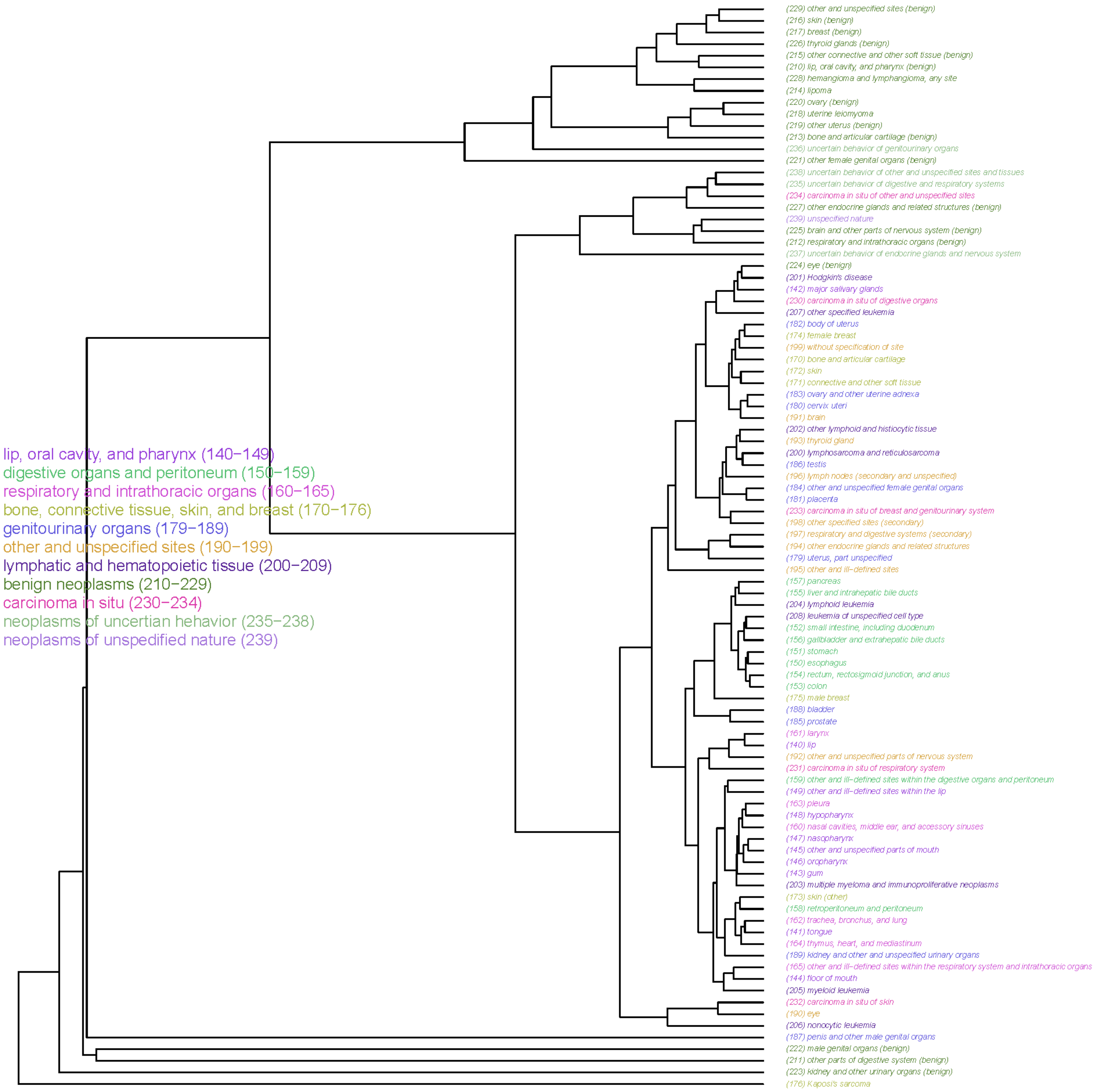

**Figure 2** Hierarchical clustering of the residual convolutional neural network classifications of herbal prescriptions into cancers. The left center legend shows categories of cancers according to International Statistical Classification of Diseases and Related Health Problems.

## IV. Discussion

The findings in Figs. 2 and S1 that distinct compositions of granulated concentrated herbal extract prescriptions were made for illnesses of different physiological derangements at different anatomic locations are consistent with previous pharmacoepidemiological studies of herbal prescriptions [6,7]. While the previous studies utilized conventional methods, i.e. frequentist statistics, on data volume of small scale, i.e. one year, the current study employs AI techniques on 10-year data of herbal prescriptions. As cancer is a relatively better studied and known disease, the result of Fig. 3 collaborates a vast body of molecular and cellular studies that tumorigenesis involves dysregulated metabolism and immunity [12]. The generalization power of AI will give it an edge in situations where novelties, e.g. new comorbidities or new co-components, are needed.

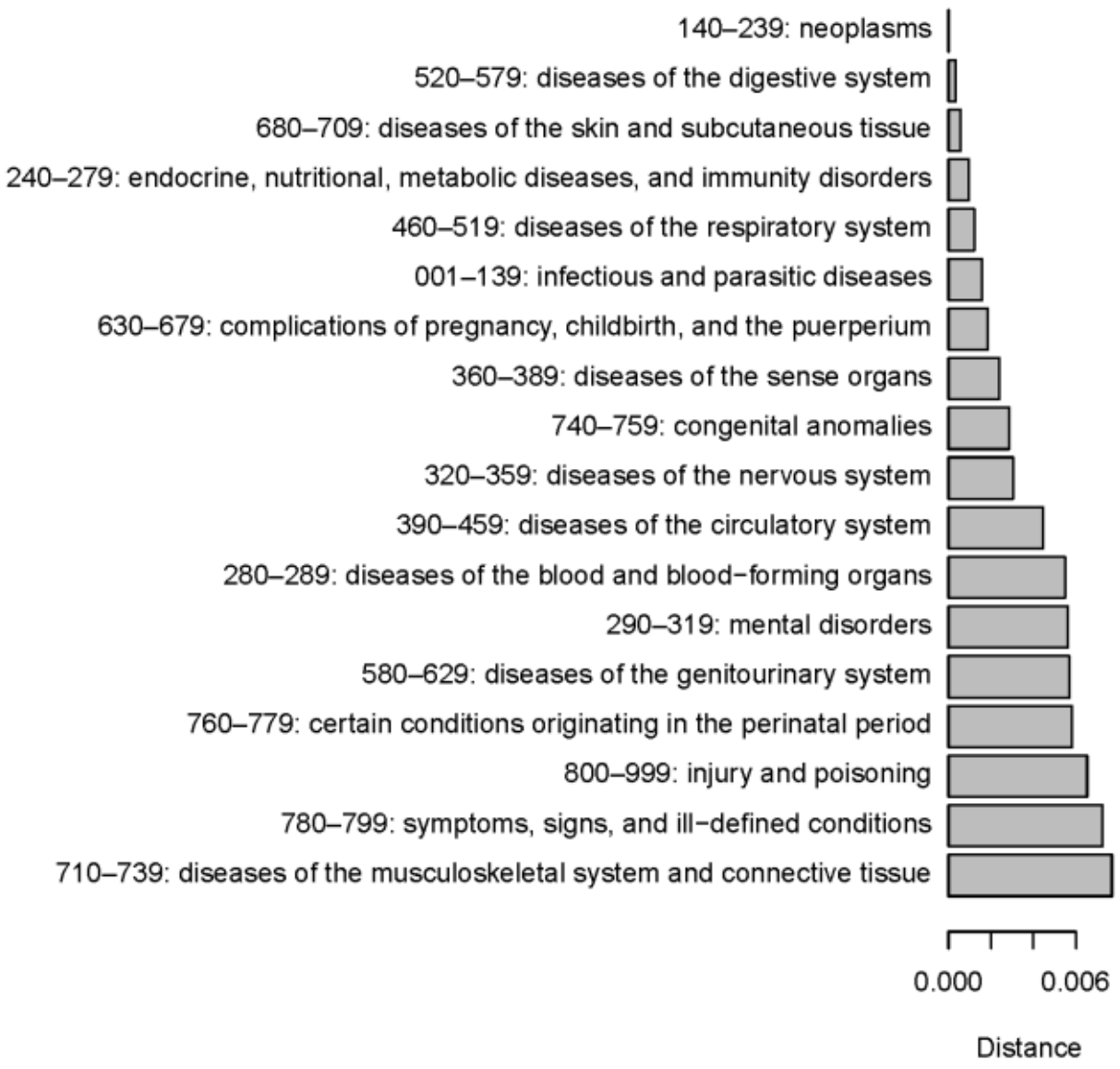

**Figure 3** Differences between cancers and other categories of diseases. Skin inflammation is closely related to immune disorder.

The current study further finds that separate herbal prescriptions were made for patients of different ages. In particular, 7-year appears to be the unit of age, which is consistent with classic TCM canons, which state that females replace deciduous teeth at 7, begin menses at 14, ready permanent teeth at 21, prime the physique at 28, start losing hairs at 35, dry up the face at 42 and stop menstruation at 49. A similar lifetime course is described for males but at a pace of 8 years. Our study also finds that herbal prescriptions grouped into seasons and years. TCM classics attribute illnesses to six exogenous – *wind*, *cold*, *warm*, *damp*, *dry*, and *fire* – and seven endogenous – *joy*, *anger*, *worry*, *reason*, *sad*, *fear* and *panic* – attrition factors. The exogenous factors are meteorological in nature and the endogenous factors are of emotional characteristics. We downloaded the monthly meteorological data, including temperature, sunshine duration, precipitation and relative humidity, of Taiwan between 2004 and 2013 [13], among the clustering of which, temperature data was found to give similar clustering pattern to Fig. S5. It suggests ambient temperature to be the major exogenous weather factor accounting for people's health in modern dwelling in Taiwan. Likewise, we retrieved the historical socioeconomic data, including gross domestic product (GDP) [14], gross national income (GNI) [14], consumer confidence index [15], Taiwan stock exchange weighted index [16], and unemployment rate of Taiwan [17] in the period from 2004 to 2013. It was found that, among them, GDP and GNI clusterings resemble Supplementary Fig. S6 [18]. Taiwan's GDP and GNI are closely related, and as they measure prosperity of the society which may be (inversely) linked to the relaxedness (stressfulness) experienced by the people of the society, GDP in modern times is likely to serve as an aggregate factor mediating population health [19].

It is challenging to compare the performance of aRNN translation as there is no phenotype-herbalism equivalent of BLEU. Nevertheless, we conducted assessment in the following way: we input aRNN output, i.e., herbal prescription, to rCNN and see if the rCNN output, i.e. phenotype, is identical to the aRNN input. That is, we tested if the two independent AI approaches generated consistent results. The results were affirmative in the examples we tested. Moreover, we generated 718 artificial herbal prescriptions, each consisting of one of the 303 formulas or 415 herbs. The rCNN predicted the class members of the artificial prescriptions with probabilities in the final softmax outputs. Clustering of the probabilities shows how similar herbal components are to one another in treating diseases, as shown in Supplementary Fig. S7. The same information is encoded in the decoder embedding of Fig. 5, whose caption shows four examples of clustered formulas/herbs in different colors. A comparison of Fig. S7 and Fig. 5 reveals agreeable clustering.

TABLE II
EXAMPLES OF TRANSLATION BY ARNN

| Case | Source sentence* | Target sentence |
|---|---|---|
| 1 | Cerebral thrombosis with cerebral infarction (43401), NA (0), NA (0) | BU3 YANG2 HUAN2 WU3 TANG1 (5.0G), XIAO3, XU4 MING4 TANG1 (2.8G), DAN1SHEN1 (2.0G), NIU2XI1 (1.5G), DU4ZHONG4 (1.3G), SANG1JI4SHENG1 (1.1G), XU4 DUAN4 (1.0G); 3 TIMES A DAY; 7 DAYS |
| 2 | Intracerebral hemorrhage (431), NA (0), NA (0) | BU3 YANG2 HUAN2 WU3 TANG1 (5.0G), TIAN1MA2 GOU1 TENG2 YIN3 (2.8G), XUE4 FU3 ZHU2 YU1 TANG1 (2.0G), SAN1QI1 (1.5G), DAN1SHEN1 (1.3G), NIU2XI1 (1.1G), DI4 LONG2 (1.0G); 3 TIMES A DAY, 7 DAYS |
| 3 | Cerebral thrombosis with cerebral infarction (43401), Atherosclerosis (440), NA (0) | BU3 YANG2 HUAN2 WU3 TANG1 (5G), XUE4 FU3 ZHU2 YU1 TANG1 (3.3G), DAN1SHEN1 (2.6G), TIAN1MA2 (2.2G), SHI2 CHANG1 PU2 (1.9G), YUAN3ZHI4 (1.7G), YU4 JIN1 (1.6G); 3 TIMES A DAY; 7 DAYS |
| 4 | Essential hypertension (401), NA (0), NA (0) | TIAN1MA2 GOU1 TENG2 YIN3 (5G); 3 TIMES A DAY; 6 DAYS |
| 5 | Atherosclerosis (440), NA (0), NA (0) | XUE4 FU3 ZHU2 YU1 TANG1 (5G), FU4FANG1 DAN1 SHEN1 PIAN4 (3.3G); 3 TIMES A DAY; 7 DAYS |

* For a 65-year old male in March, year not specified; diseases are in primary, secondary and tertiary order.

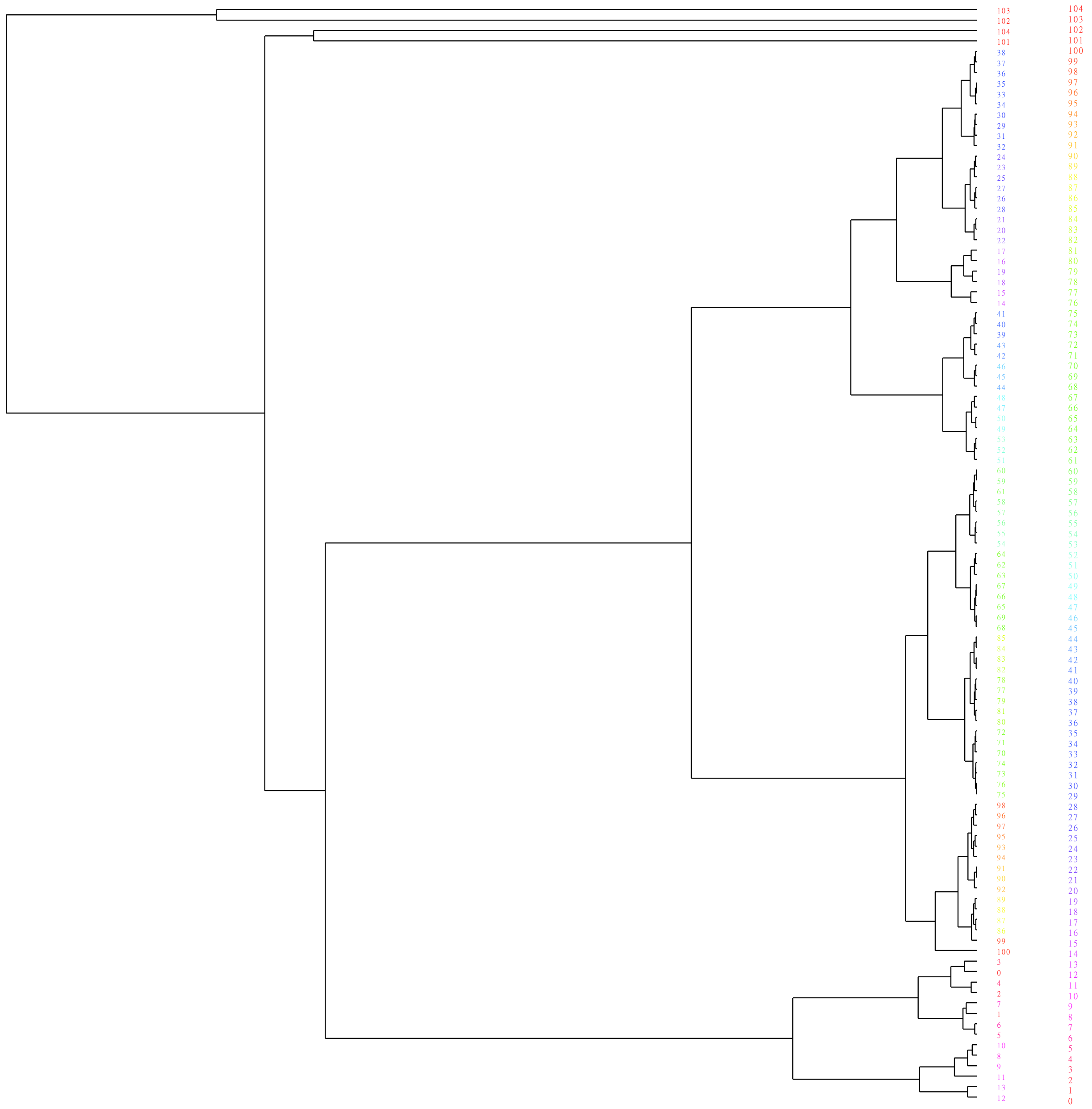

**Figure 4** Hierarchical clustering of the residual convolutional neural network classifications of herbal prescriptions into patient ages. The right bar colors ages.

Latest studies suggested that neural machines learned not only syntax but also semantics of natural languages [20]. Stroke is known to be due to lack of blood or bleeding in the brain. The former, called ischemic stroke, has two subtypes (ICD-9 codes): cerebral thrombosis (43401) and cerebral embolism (43411); the latter, called hemorrhagic stroke, has also two subtypes: cerebral hemorrhage (431) and subarachnoid hemorrhage (430). Ischemic stroke is more common (~85%) and less fatal. However, both strokes can have similar functional outcomes as the ruptured artery in hemorrhagic stroke ceases feeding blood to the parts beyond the rupture. Table 2 shows the output prescriptions by aRNN for a 65-year-old male with thrombotic stroke (ICD-9 = 43401, 0, 0), hemorrhagic stroke (ICD-9 = 431, 0, 0) or thrombotic stroke with atherosclerosis (ICD-9 = 43401, 440, 0). In classic TCM formula composing, the heaviest herbs, called *Monarch*, are meant to antagonize the primary symptom, the second heaviest herbs, called *Minister*, antagonize secondary symptom. The next heaviest herbs, called *Assistant*, complement *Monarch* and *Minister*, and the lightest herbs, called *Guide*, guide the whole formula to the targeted TCM meridian. The examples in Table 2 show that, in modern herbal prescription composing using granulated concentrated herbal extracts, the heaviest components in the prescription are granulated classic TCM formulas and lighter components are granulated TCM herbs. The fact that the herbs are usually components in the formulas suggests that the herbs be meant to fine-tune effects of the granulated classic TCM formulas. In other words, in classic formulating, phrases are made from words; in modern formulating, sentences are composed out of phrases and words. The aRNN has therefore, in the sense of component ordering (i.e. weighting), learned the syntax of herbal prescribing.

Let's examine Table 2 in further detail, focusing on the differences first. It is known in western medicine that atherosclerotic plaques directly cause ischemic stroke. It is also known that hemorrhagic stroke can arise from hypertension, which is in turn associated with atherosclerosis. Tian Ma Gou Teng Yin (TMGTY) alone constitutes a prescription for hypertension as shown in case 4 of Table 2. This explains the appearance of TMGTY in the hemorrhagic stroke prescription in the second case in Table 2. Xue Fu Zhu Yu Tang (XFZYT) is the dominant formula in the prescription for atherosclerosis as shown in case 5 of Table 2. This explains the appearance of XFZYT in the second case. Furthermore, as atherosclerosis is included as the secondary disease to the thrombotic stroke patient, XFZYT shows up in the prescription in the third case, in contrast to the prescription for the patient with thrombotic stroke alone in the first case in Table 2. The decreasing doses, i.e. the order, of the formulas in the aRNN prescriptions reflect: 1) the relative distance between the associated risk factors and the disease in the case of hemorrhagic stroke; and 2) the primary and secondary comorbidity in the case of thrombotic stroke with atherosclerosis. It is also of note that if the disease is more deadly or if the patient has more than one disease, the Zipf's exponent, *z*, is less, resulting in a slower decay of the component weights. Altogether, with our equivalating phenotypes to English, and prescriptions to French, it is argued that aRNN has learnt the semantics of both languages and are translating phenotypes to herbal prescriptions.

Finally, let's address the commonalities, that is, Bu Yang Huan Wu Tang (BYHWT) and those single herbs in the three stroke prescriptions. A recent study used: 1) compound database for traditional Chinese medicinal plants; 2) known anti-stroke molecular targets; and 3) molecular docking between compound and target to screen for anti-stroke compounds [21]. The result identified 192 anti-stroke plants, the top plants being Chuan Xiong, Dang Gui, Hong Hua, Chi Shao, Huang Qi, Dan Shen, Tao Ren, Da Huang, and Chang Pu. BYHWT is known to be made up of Huang Qi, Dang Gui Wei, Chi Shao, Di Long, Chuan Xiong, Tao Ren and Hong Hua [22]. Therefore, BYHWT and the single herbs in the above aRNN prescriptions encompass those anti-stroke plants recently identified using modern biotechnologies and bioinformatics. As the quality of AI predictions is commensurate with that of the training data, it is clear that we should credit both herbalist doctors for their prescribing and also modern western medicine doctors for their diagnosing using modern medical devices.

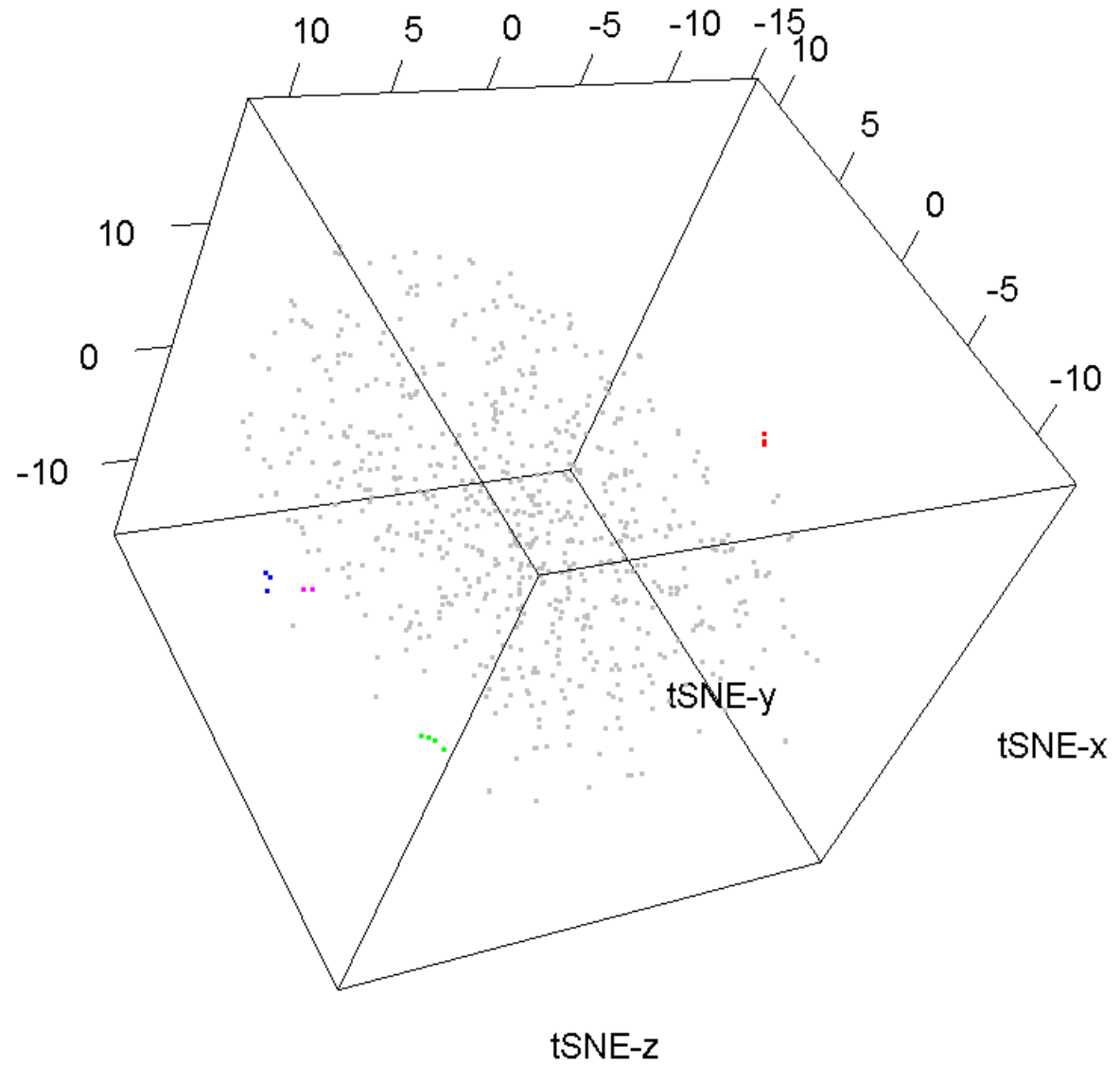

**Figure 5** A three-dimensional tSNE projection of the attenstion-based recurrent neural network decoder embeddings. The three blue points are Tao2 He2 Cheng2 Qi4 Tang1, Diao4 Wei4 Cheng2 Qi4 Tang1 and Di3 Dang1 Tang1. The two magenta points are Da4 Cheng2 Qi4 Tang1 and Xiao3 Cheng2 Qi4 Tang1. The four green points are Qing1 Shang4 Fang2 Feng1 Tang1, Yi3 Zi4 Tang1, Wan2 Dai4 Tang1 and Ba1 Wei4 Dai4 Xia4 Fang1. The three red points are Gui1 Lu4 Er4 Xian1 Jiao1, Hu3 Qian2 Wan2 (excluding Hu3Gu3) and Hong2 Qu1. Color caption online.

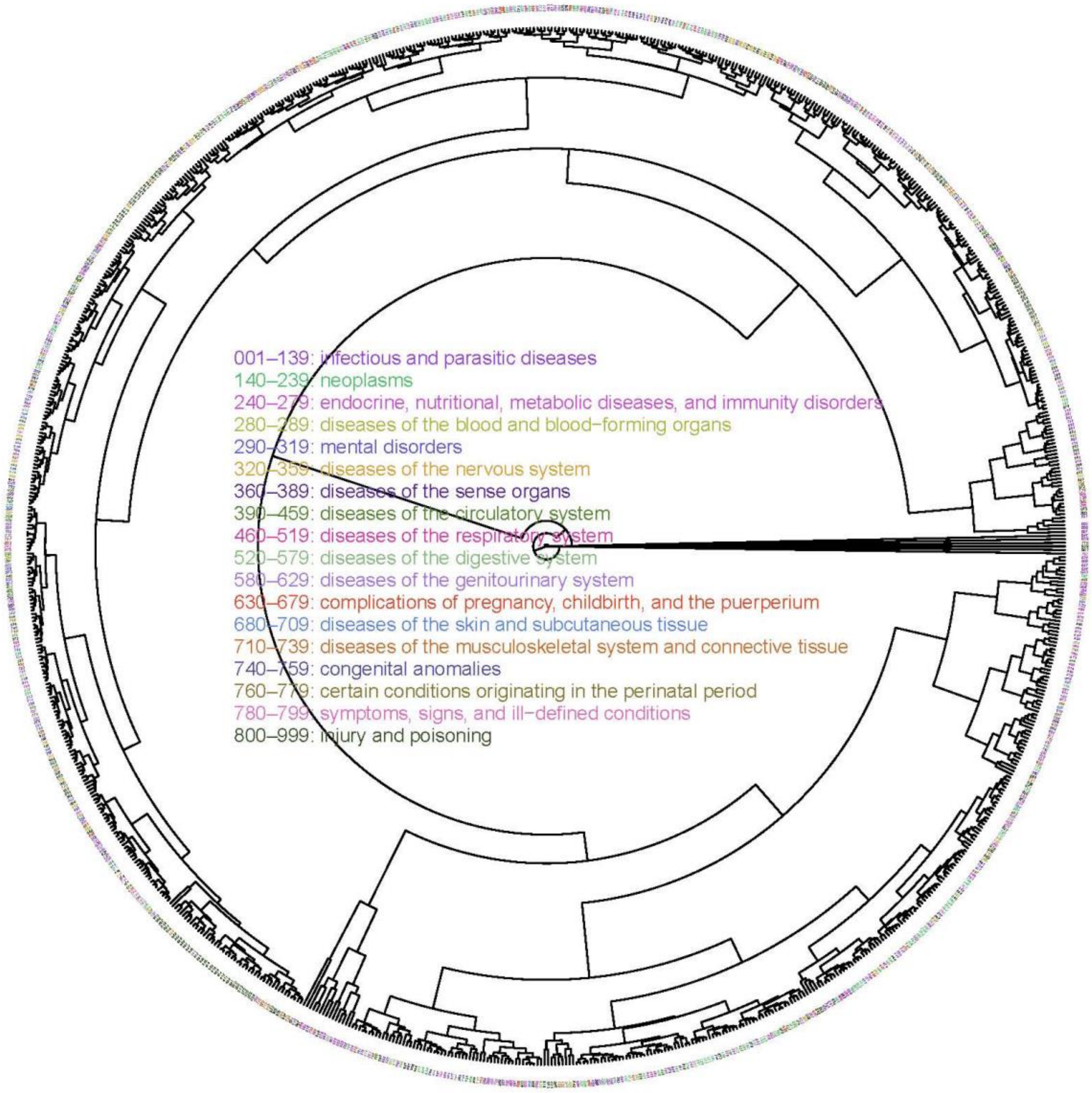

**Figure S1** Hierarchical clustering of residual convolutional neural network classifications of herbal prescriptions into diseases.

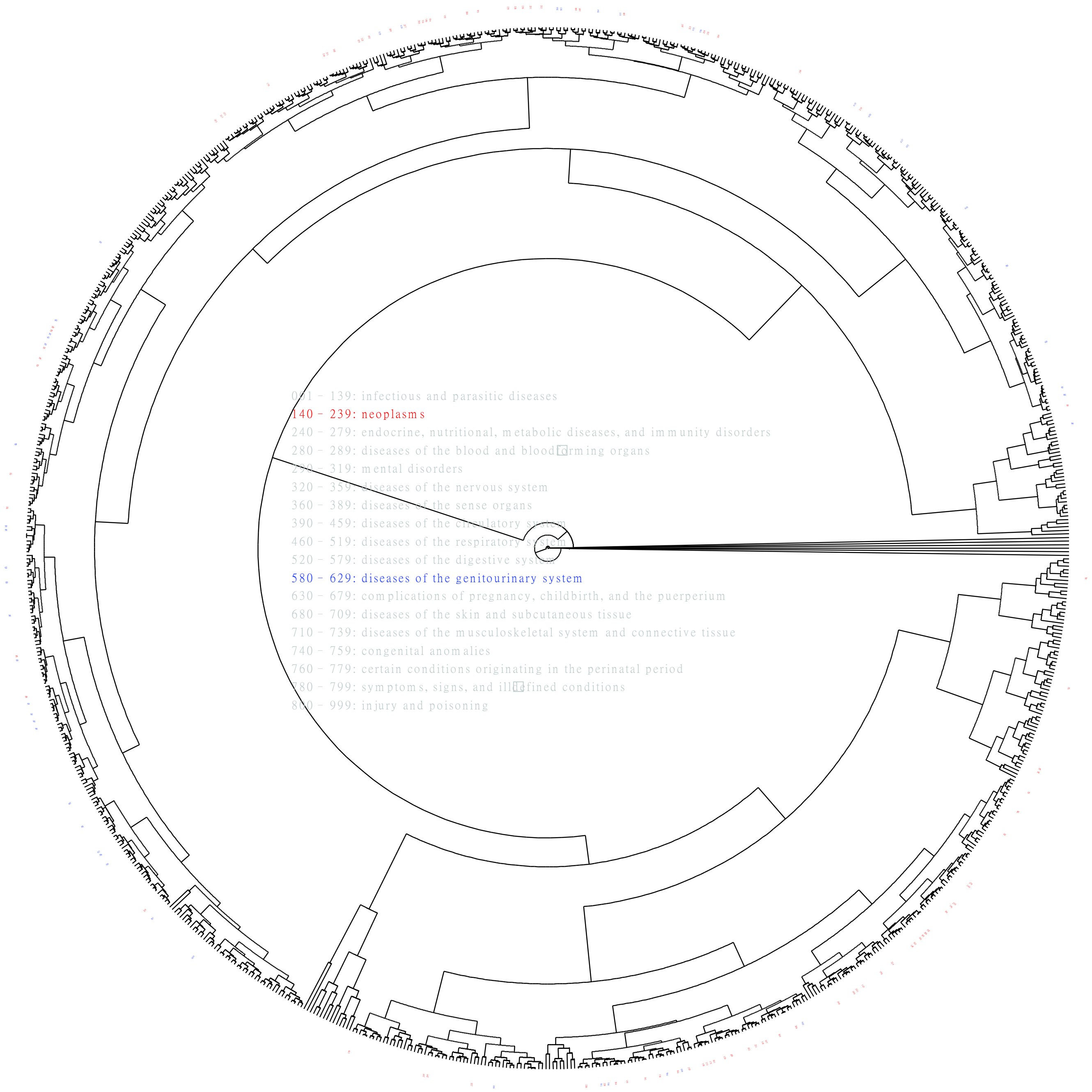

**Figure S2** Hierarchical clustering of genitourinary system diseases.

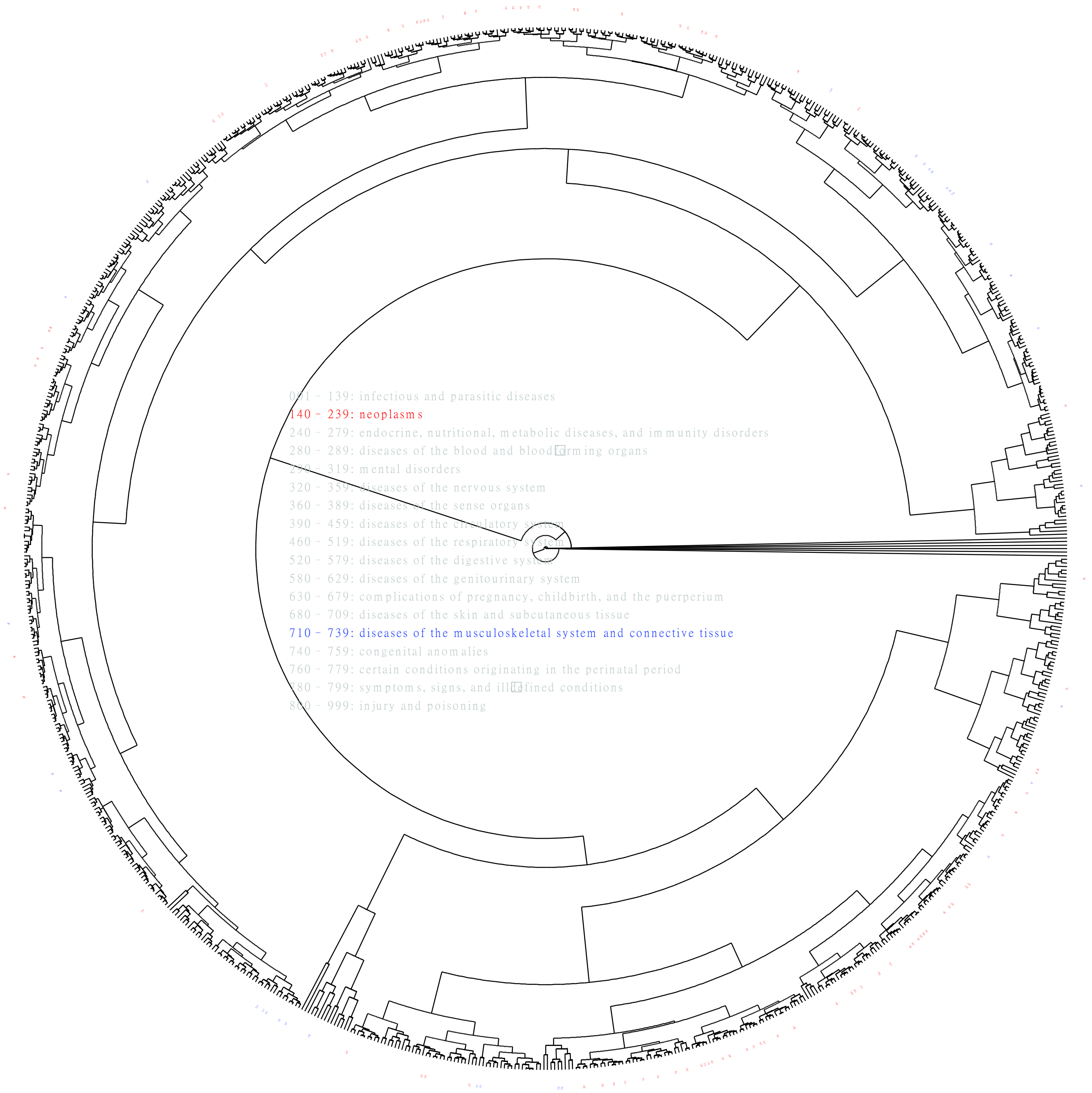

**Figure S3** Hierarchical clustering of musculoskeletal system and connective tissue diseases.

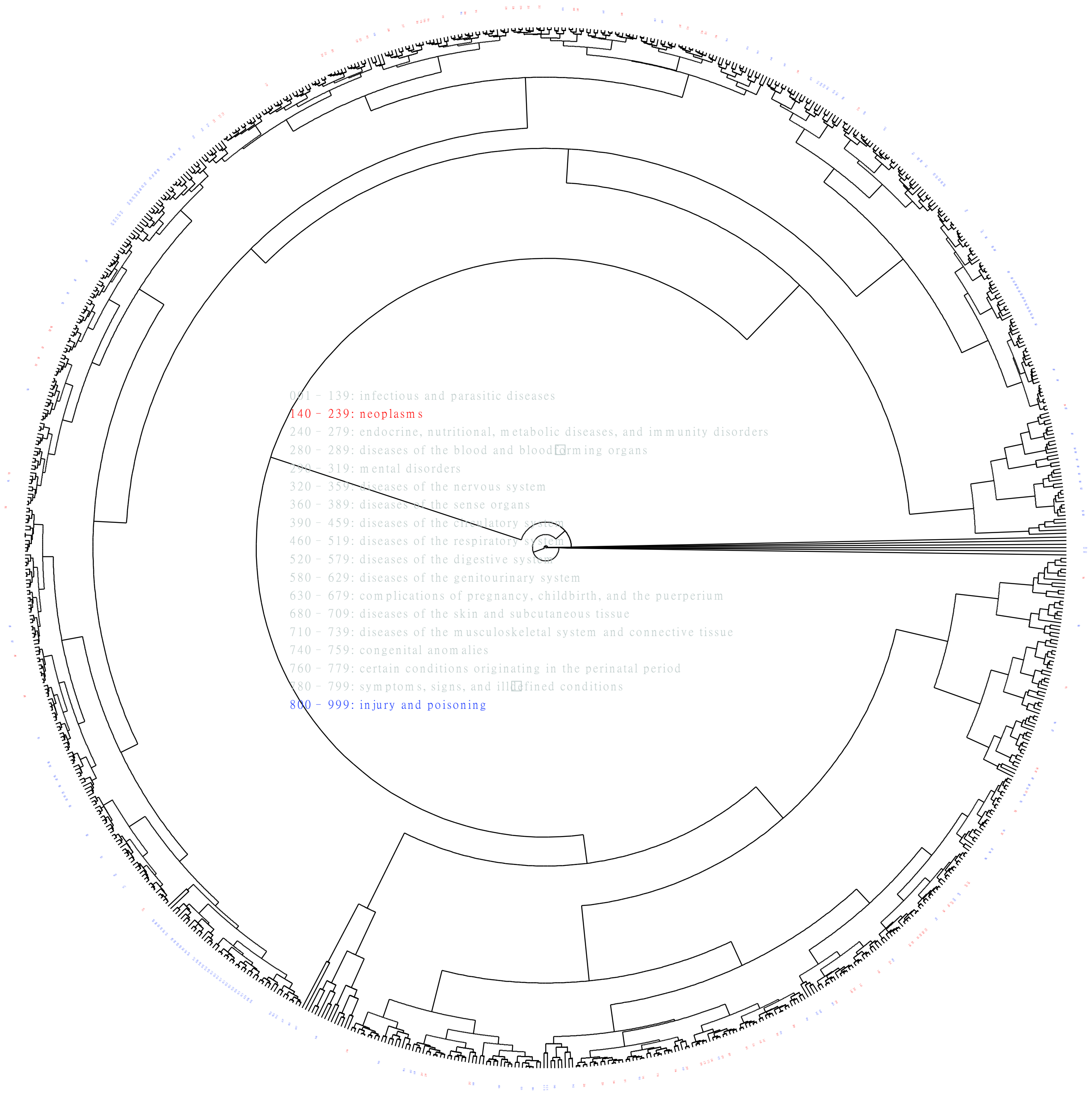

**Figure S4** Hierarchical clustering of injury and poisoning.

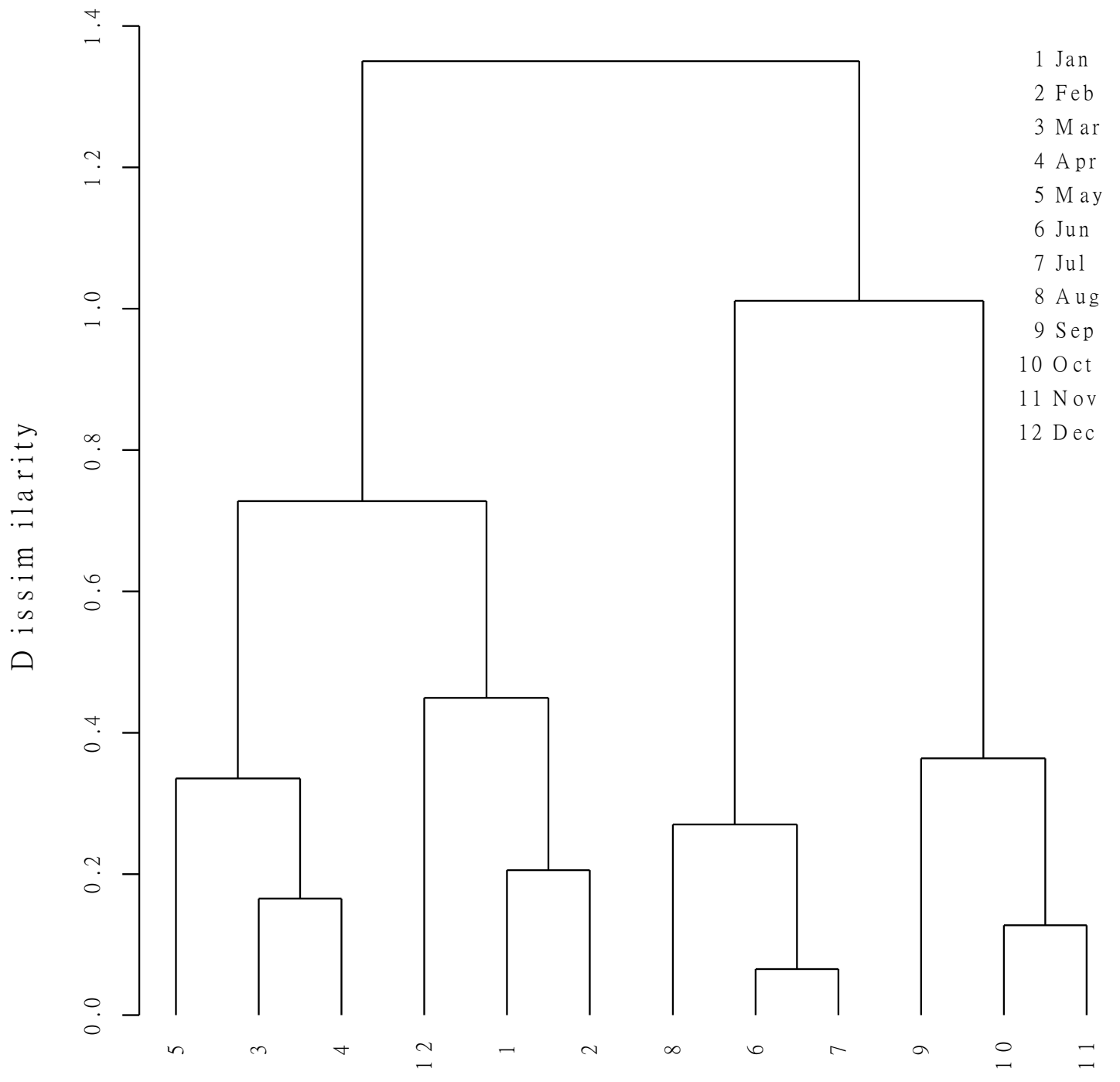

**Figure S5** Hierarchical clustering of residual convolutional neural network classifications of herbal prescriptions into months.

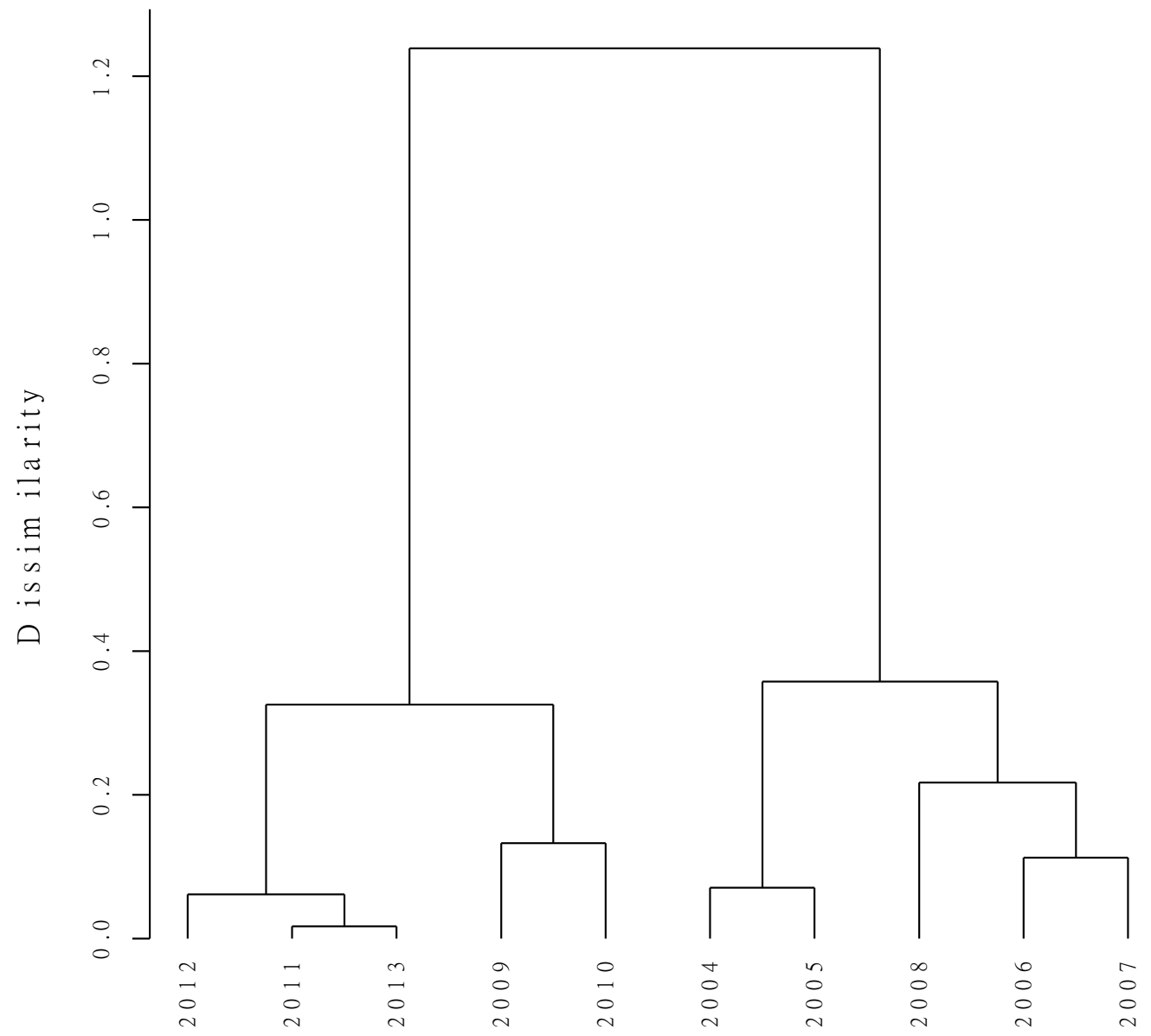

**Figure S6** Hierarchical clustering of residual convolutional neural network classifications of herbal prescriptions into years.

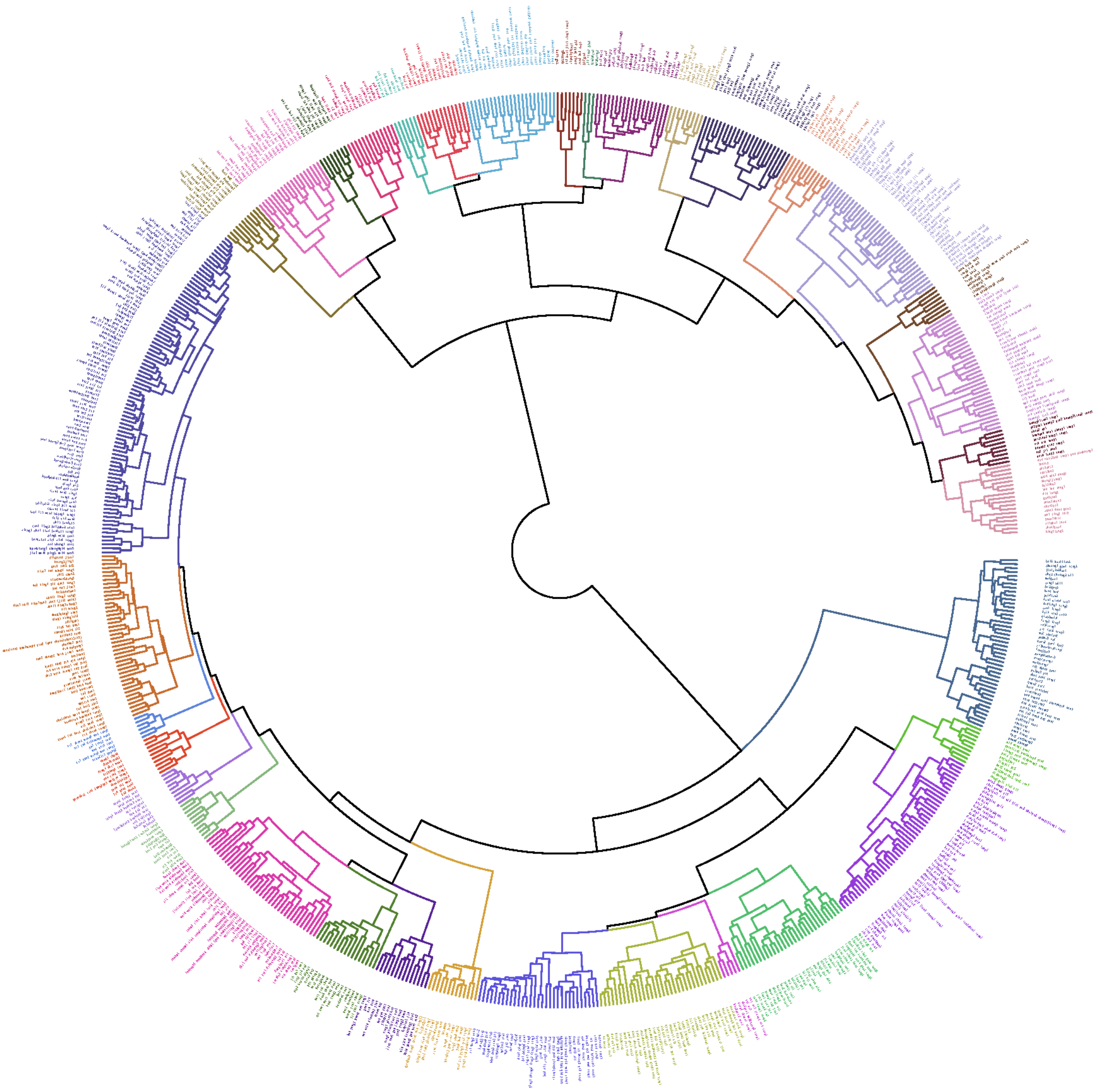

**Figure S7** Hierarchical clustering of residual convolutional neural network classifications of herbal prescriptions.